\documentclass[preprint,12pt]{elsarticle}%
\usepackage{graphicx}
\usepackage{float}
\usepackage{url}
\usepackage{amsmath}
\usepackage{amsfonts}
\usepackage{amssymb}
\usepackage{geometry}%
\usepackage[ruled,vlined]{algorithm2e}
\usepackage{comment}
\providecommand{\U}[1]{\protect\rule{.1in}{.1in}}
\begin{document}
%


\begin{frontmatter}

\title{A Neural Network-Evolutionary Computational Framework for Remaining Useful Life Estimation of Mechanical Systems}

\author{David Laredo$^{1}$, Zhaoyin Chen$^{1}$, Oliver Sch\"utze$^{2}$ and Jian-Qiao Sun$^{1}$}
\address{
$^{1}$Department of Mechanical Engineering\\
School of Engineering, University of California\\
Merced, CA 95343, USA\\
$^{2}$Department of Computer Science, CINVESTAV\\ 
Mexico City, Mexico\\
Corresponding author. Email: davidlaredo1@gmail.com}

\begin{abstract}
This paper presents a framework for estimating the remaining useful life (RUL) of mechanical systems. The framework consists of a multi-layer perceptron and an evolutionary algorithm for optimizing the data-related parameters. The framework makes use of a strided time window to estimate the RUL for mechanical components. Tuning the data-related parameters can become a very time consuming task. The framework presented here automatically reshapes the data such that the efficiency of the model is increased. Furthermore, the complexity of the model is kept low, e.g. neural networks with few hidden layers and few neurons at each layer. Having simple models has several advantages like short training times and the capacity of being in environments with limited computational resources such as embedded systems. The proposed method is evaluated on the publicly available C-MAPSS dataset \cite{Saxena2008a}, its accuracy is compared against other state-of-the art methods for the same dataset.

\end{abstract}


\end{frontmatter}



\section{Introduction}

\label{sec:rul_intro}

Traditionally, maintenance of mechanical systems has been carried out based on
scheduling strategies. Such strategies are often costly and less capable of
meeting the increasing demand of efficiency and reliability
\cite{Gebraeel2005, Zaidan2013}. Condition based maintenance (CBM) also known
as intelligent prognostics and health management (PHM) allows for maintenance
based on the current health of the system, thus cutting down the costs and increasing
the reliability of the system \cite{Zhao2017}. Here, we refer to prognostics
as the estimation of remaining useful life of a system. The remaining useful life (RUL) of
the system can be estimated based on the historical data. This data-driven
approach can help optimize maintenance schedules to avoid engineering failures
and to save costs \cite{Lee2014}.

The existing PHM methods can be grouped into three different categories:
model-based \cite{Yu2001}, data-driven \cite{Liu2009, Mosallam2013} and
hybrid approaches \cite{Pecht2010, Liu2012}. Model-based approaches attempt to
incorporate physical models of the system into the estimation of the RUL. If
the system degradation is modeled precisely, model-based approaches usually
exhibit better performance than data-driven approaches \cite{Qian2017}. This
comes at the expense of having extensive a priori knowledge of the underlying
system and having a fine-grained model of the system, which can involve
expensive computations. On the other hand, data-driven approaches use pattern
recognition to detect changes in system states. Data-driven approaches are
appropriate when the understanding of the first principles of the system
dynamics is not comprehensive or when the system is sufficiently complex such as
jet engines, car engines and complex machineries, for which it is prohibitively difficult 
to develop an accurate model.

Common disadvantages for the data-driven approaches are that they usually
exhibit wider confidence intervals than model-based approaches and that a fair
amount of data is required for training. Many data-driven algorithms have been
proposed. Good prognostics results have been achieved. Among the most popular
algorithms we can find artificial neural networks (ANNs) \cite{Gebraeel2005},
support vector machine (SVM) \cite{Benkedjouh2013}, Markov hidden chains (MHC)
\cite{Dong2007} and so on. Over the past few years, data-driven approaches
have gained more attention in the PHM community. A number of machine learning
techniques, especially neural networks, have been applied successfully to
estimate the RUL of diverse mechanical systems. ANNs have demonstrated good
performance in modeling highly nonlinear, complex, multi-dimensional systems
without any prior knowledge on the system behavior \cite{Li2018}. While the
confidence limits for the RUL predictions cannot be analytically provided
\cite{Sikorska2011}, the neural network approaches are promising for
prognostic problems.

Neural networks for estimating the RUL of jet engines have been previously
explored in \cite{Lim2016} where the authors propose a multi-layer perceptron
(MLP) coupled with a feature extraction (FE) method and a time window for the
generation of the features for the MLP. In the publication, the authors
demonstrate that a moving window combined with a suitable feature extractor
can improve the RUL prediction as compared with the studies with other similar
methods in the literature. In \cite{Li2018}, the authors explore a deep
learning ANN architecture, the so-called convolutional neural networks (CNNs),
where they demonstrate that by using a CNN without any pooling layers coupled
with a time window, the predicted RUL is further improved.

In this paper we propose a novel framework for estimating the RUL of complex
mechanical systems. The framework consists of a MLP to estimate the RUL of the
system, coupled with an evolutionary algorithm for the fine tuning of
data-related parameters. In this paper we refer to data-related parameters as parameters that define the shape, defined in terms of window size and window stride, and quality of the data, measured with respect to some performance indicators, used by the MLP. Please note that while this specific framework makes use of a MLP, 
the framework can in principle use several other learning algorithms, our main objective is the 
treatment of the data instead of the choice of a particular learning algorithm.
The publicly available NASA C-MAPSS dataset
\cite{Saxena2008a} is used to assess the efficiency and reliability of the
proposed framework, by efficiency we mean the complexity of the used regressor and reliability refers to the accuracy of the predictions made by the regressor. This approach allows for a simple and small MLP to obtain
better results than those reported in the current literature while using less
computing power.

The remainder of this paper is organized as follows. The C-MAPSS dataset is
presented in Section \ref{sec:rul_dataset}. The framework and its components
are thoroughly reviewed in Section \ref{sec:method}. The method is evaluated
using the C-MAPSS dataset in Section \ref{sec:rul_eval}. A comparison with the
state-of-the-art is also provided. Finally, the conclusions are presented in
Section \ref{sec:conclusions}.

\section{NASA C-MAPSS Dataset}
\label{sec:rul_dataset}

The NASA C-MAPSS dataset is used to evaluate performance of the proposed method \cite{Saxena2008a}. The C-MAPSS dataset contains simulated data produced using a model based simulation program developed by NASA. The dataset is further divided into 4 subsets composed of multi-variate temporal data obtained from 21 sensors.

For each of the 4 subsets, a training and a test set are provided. The training sets include run-to-failure sensor records of multiple aero-engines collected under different operational conditions and fault modes as described in Table \ref{table:CMAPSS}.

\begin{table}[H]
\begin{center}
\begin{tabular}
[c]{l|cccc}\hline
& \multicolumn{4}{c}{C-MAPSS}\\
Dataset & FD001 & FD002 & FD003 & FD004\\ \hline
Training Trajectories & 100 & 260 & 100 & 248\\
Test Trajectories & 100 & 259 & 100 & 248\\
Operating Conditions & 1 & 6 & 1 & 6\\
Fault Modes & 1 & 1 & 2 & 2\\\hline
\end{tabular}
\caption{C-MAPSS dataset details.}
\label{table:CMAPSS}
\end{center}
\end{table}

The data is arranged in an $N\times26$ matrix where $N$ is the number of data points in each subset. The first two variables represent the engine and cycle numbers, respectively. The following three variables are operational settings which correspond to the conditions in Table \ref{table:CMAPSS} and have a substantial effect on the engine performance. The remaining variables represent the 21 sensor readings that contain the information about the engine degradation over time.

Each trajectory within the training and test sets represents the life cycles of the engine. Each engine is simulated with different initial health conditions, i.e. no initial faults. For each trajectory of an engine the last data entry corresponds to the cycle at which the engine is found faulty. On the other hand, the trajectories of the test sets terminate at some point prior to failure, hence the need to predict the remaining useful life. The aim of the MLP model is to predict the RUL of each engine in the test set. The actual RUL values of test trajectories are also included in the dataset for verification. Further discussions of the dataset and details on how the data is generated can be found in \cite{Saxena2008}.

\subsection{Performance Metrics}

\label{sec:rul_metrics}

To evaluate the performance of the proposed approach on the C-MAPSS dataset, we make use of two scoring indicators, namely the Root Mean Squared Error (RMSE) denoted as $E_{\scriptscriptstyle RMS}(\mathbf{e})$ and a score proposed in \cite{Saxena2008} which we refer as the RUL Health Score (RHS) denoted as $E_{\scriptscriptstyle RH}(\mathbf{e})$. The two scores are defined as follows,

\begin{equation}
E_{\scriptscriptstyle RMS} = \sqrt{ \frac{1}{n} \sum_{i=1}^{N}{e_{i}^{2}}} \label{eq:rmse}%
\end{equation}

\begin{align}
E_{\scriptscriptstyle RH}  &  = \frac{1}{n} \sum_{i=1}^{N}{s_{i}}\nonumber\\
s_{i}  &  =
\begin{cases}
\exp(-\frac{e_{i}}{13}) - 1, & e_{i} < 0\\
\exp(\frac{e_{i}}{10}) - 1, & e_{i} \geq0,
\end{cases}
\label{eq:rhs}%
\end{align}

where $n$ is the total number of samples in the test set and $\mathbf{e} = \mathbf{\hat{y}} - \mathbf{y}$ is the error between the estimated RUL values $\mathbf{\hat{y}}$, and the actual RUL values $\mathbf{y}$ for each engine within the test set. It is important to note that $E_{\scriptscriptstyle RH}(\mathbf{e})$ penalizes late predictions more than early predictions since usually late predictions lead to more severe consequences in fields such as aerospace.

\section{Framework Description}

\label{sec:method}

n this section, the proposed ANN-EA based method for prognostics is presented. The method makes use of a multi-layer perceptron (MLP) as the main regressor for estimating the RUL of the engines in the C-MAPSS dataset. The choice of a MLP as the learning algorithm instead of any of the other choices (SVM, RNN, CNN, Least-Squares, etc) obeys to the fact that MLPs are in general good for nonlinear data like the one exhibited by the C-MAPSS dataset, but at the same time are less computationally expensive than some of the more sophisticated algorithms as the CNN or the RNN. Indeed, the RNN may be a more suitable choice for this particular problem since it involves time-sequenced data, nevertheless, we will show that by doing a fine tuning of the data-related parameters (and thus data processing), the inference power of a simple MLP can be competitive even when compared against that of an RNN.   For the training sets, the feature vectors are generated by using a moving time window while a label vector is generated with the RUL of the engine. The label has a constant RUL for the early cycles of the simulation, and becomes a linearly decreasing function of the cycle in the remaining cycles. This is the so-called piece-wise linear degradation model \cite{Ramasso2014}. For the test set, a time window is taken from the last sensor readings of the engine. The data of the test set is used to predict the RUL of the engine.

The window-size $n_{w}$, window-stride $n_{s}$, and early-RUL $R_{e}$ are data-related parameters, which for the sake of clarity and formalism in this study, form a vector 
$\mathbf{v} \in \mathbb{Z}^{3}$ such that $\mathbf{v} = (n_{w}, n_{s}, R_{e})$. The vector $\mathbf{v}$ has a considerable impact on the quality of the predictions by the regressor. It is computationally intensive to find the best parameters of $\mathbf{v}$ given the search space inherent to these parameters. We propose an evolutionary algorithm to optimize the data-related parameters $\mathbf{v}$. The optimized parameter set $\mathbf{v}$ allows the use of a simple neural network architecture while attaining better results in terms of the quality of the predictions compared with the results by other methods in the literature.

\subsection{The Network Architecture}

After careful examinations of the C-MAPSS dataset, we propose to use a rather simple MLP architecture for all the four subsets of the data. The choice of a simple architecture over a more complex one follows the fact that simpler neural networks are less computationally expensive to train, furthermore, the inference is done also faster since less operations are involved. To measure the simplicity/complexity of a neural network we use the number of trainable parameters (weights) of the neural network, usually the more trainable parameters in a network the more computations that need be done, thus increasing the computational burden of the training/inference process. The implementations are done in Python using the Keras/Tensorflow environment. The source code is publicly available at the git repository \url{https://github.com/dlaredo/NASA_RUL_-CMAPS-} \cite{Laredo2018}.

The choice of the network architecture is made by following an iterative process, our goal was to find a good compromise between efficiency (simple neural network models) and reliability (scores obtained by the model using the RMSE metric): We compared 6 different architectures (see Appendix \ref{appendix:tested_architectures}), training each for $100$ epochs using a mini-batch size of $512$ and averaging their results on a cross-validation set for $10$ different runs. L1 (Lasso) and L2 regularization (Ridge) \cite{Buhlmann2011} are used to prevent over-fitting. L1 regularization penalizes the sum of the absolute value of the weights and biases of the networks, while L2 regularization penalizes the sum of the squared value of the weights and biases. The data-related parameters $\mathbf{v}$ used for this experiment are $\mathbf{v}= (30, 1, 140)$. Two objectives are pursued during the iterations: the architecture must be minimal in terms of the number of trainable parameters and the performance indicators must be minimized. Table \ref{table:tested_architectures_100} summarizes the results for each tested architecture.

\begin{table}[H]
\begin{center}
\begin{tabular}
[c]{l|cccc|cccc}\hline & \multicolumn{4}{|c}{RMSE} & \multicolumn{4}{|c}{RHS}\\
Tested Architecture & Min. & Max. & Avg. & STD & Min. & Max. & Avg. & STD\\\hline
Architecture 1 & 15.51 & 17.15 & 16.22 & 0.49 & 4.60 & 7.66 & 5.98 & 0.91\\
Architecture 2 & 15.24 & 16.46 & 15.87 & 0.47 & 4.07 & 6.26 & 5.29 & 0.82\\
Architecture 3 & 15.77 & 17.27 & 16.15 & 0.45 & 5.11 & 8.25 & 5.93 & 0.94\\
Architecture 4 & 15.13 & 17.01 & 15.97 & 0.47 & 3.90 & 7.54 & 5.65 & 1.2\\
Architecture 5 & 16.39 & 17.14 & 16.81 & 0.23 & 5.19 & 6.58 & 5.98 & 0.42\\
Architecture 6 & 16.42 & 17.36 & 16.87 & 0.30 & 5.15 & 7.09 & 6.12 & 0.62\\\hline
\end{tabular}
\caption{Results for different architectures for subset 1, 100 epochs.}
\label{table:tested_architectures_100}
\end{center}
\end{table}

Table \ref{table:proposed_nn} presents the architecture chosen for the remainder of this work (which provides the best compromise between compactness and performance among the tested architectures). Each row in the table represents a neural network layer while each column describes each one of the key parameters of the layer such as the type of layer, number of neurons in the layer, activation function of the layer and whether regularization is used, where L1 denotes the L1 regularization factor and L2 denotes the L2 regularization factor, the order in which the layers are appended from the table is top-bottom. From here on we refer to this neural network model as $\phi(.)$.

\begin{table}[H]
\begin{center}
\begin{tabular}
[c]{llll}\hline
Layer & Neurons & Activation & Additional Information\\\hline
Fully connected & \multicolumn{1}{c}{20} & \multicolumn{1}{c}{ReLU} &
$L1=0.1,L2=0.2$\\
Fully connected & \multicolumn{1}{c}{20} & \multicolumn{1}{c}{ReLU} &
$L1=0.1,L2=0.2$\\
Fully connected & \multicolumn{1}{c}{1} & \multicolumn{1}{c}{Linear} &
$L1=0.1,L2=0.2$\\\hline
\end{tabular}
\caption{Proposed neural network architecture $\phi(.)$.}
\label{table:proposed_nn}
\end{center}
\end{table}

\subsection{Shaping the Data}

This section covers the data pre-processing applied to the raw sensor readings in each of the datasets. Although the original datasets contain $21$ different sensor readings, some of the sensors do not present much variance or convey redundant information, our choice of the sensors is based on previous studies such as \cite{Lim2016, Li2018} where it was discovered that some sensor values do not vary at all throughout the entire engine life cycle while some others are redundant according to PCA or clustering analysis. These sensors are therefore discarded. In the end, only $14$ sensor readings out of the $21$ are considered for this study. Their indices are $\left\lbrace 2, 3, 4, 7, 8, 9, 11, 12, 13, 14, 15, 17, 20, 21 \right\rbrace $. The raw measurements are then used to create the strided time windows with window-size $n_{w}$ and window-stride $n_{s}$. For the training labels, $R_{e}$ is used at the early stages and then the RUL is linearly decreased. Assuming $\mathbf{x} \in \mathbb{R}^m$ is the vector whose components are the sensor readings at each time stamp, then the min-max normalized vector $\mathbf{\hat{x}}$ can be computed by means of the following formula: 

\begin{equation}
\hat{x}_{i} = 2* \frac{x_{i} - min(x_{i})}{max(x_{i}) - min(x_{i})} - 1.
\label{eq:min_max_norm}%
\end{equation}

\subsubsection{Time Window and Stride}

\begin{figure}[H]
\begin{center}
\includegraphics[scale=0.2]{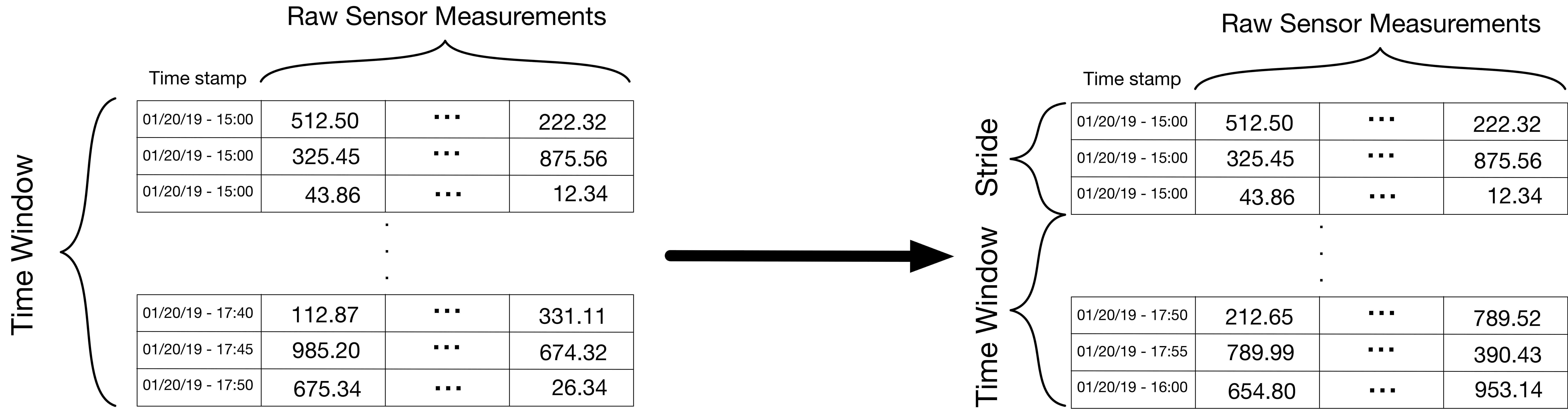}
\caption{Graphical depiction of the time window used in this framework.}
\label{fig:moving_window}
\end{center}
\end{figure}

In multivariate time-series problems such as RUL, more information can be generally obtained from the temporal sequence of the data as compared with the multivariate data point at a single time stamp. For a time window of size $n_{w}$ with a stride $n_{s}=1$, all the sensor readings in the time window form a feature vector $\mathbf{x} \in \mathbb{R}^{s*n_w}$, where $s$ denotes the number of sensors being read. Stacking together $m_w$ of this time windows forms feature vector $\mathbf{X} \in \mathbb{R}^{m_w \times s*n_w}$ while its corresponding RUL values are defined as $\mathbf{y} \in \mathbb{Z}^m$. It is important to mention that the shape of $\mathbf{X}$ defines the number of input neurons for the neural network, therefore changing the shape of $\mathbf{X}$ effectively changes the number of inputs to the neural network. This approach has successfully been tested in \cite{Li2018,Lim2016} where the authors propose the use of a moving window with sizes ranging from 20 to 30. We propose not only the use of a moving time window, but also a \textit{strided} time window that updates more than one element ($n_s>1$) at the time. A graphical depiction of the strided time window is shown in Figure \ref{fig:moving_window}. For Figure \ref{fig:moving_window} the numbers and time-stamps are just illustrative, the window size exemplified is of 30 time-stamps while the stride is of 3 time-stamps.

\begin{figure}[H]
\begin{center}
\includegraphics[scale=0.8]{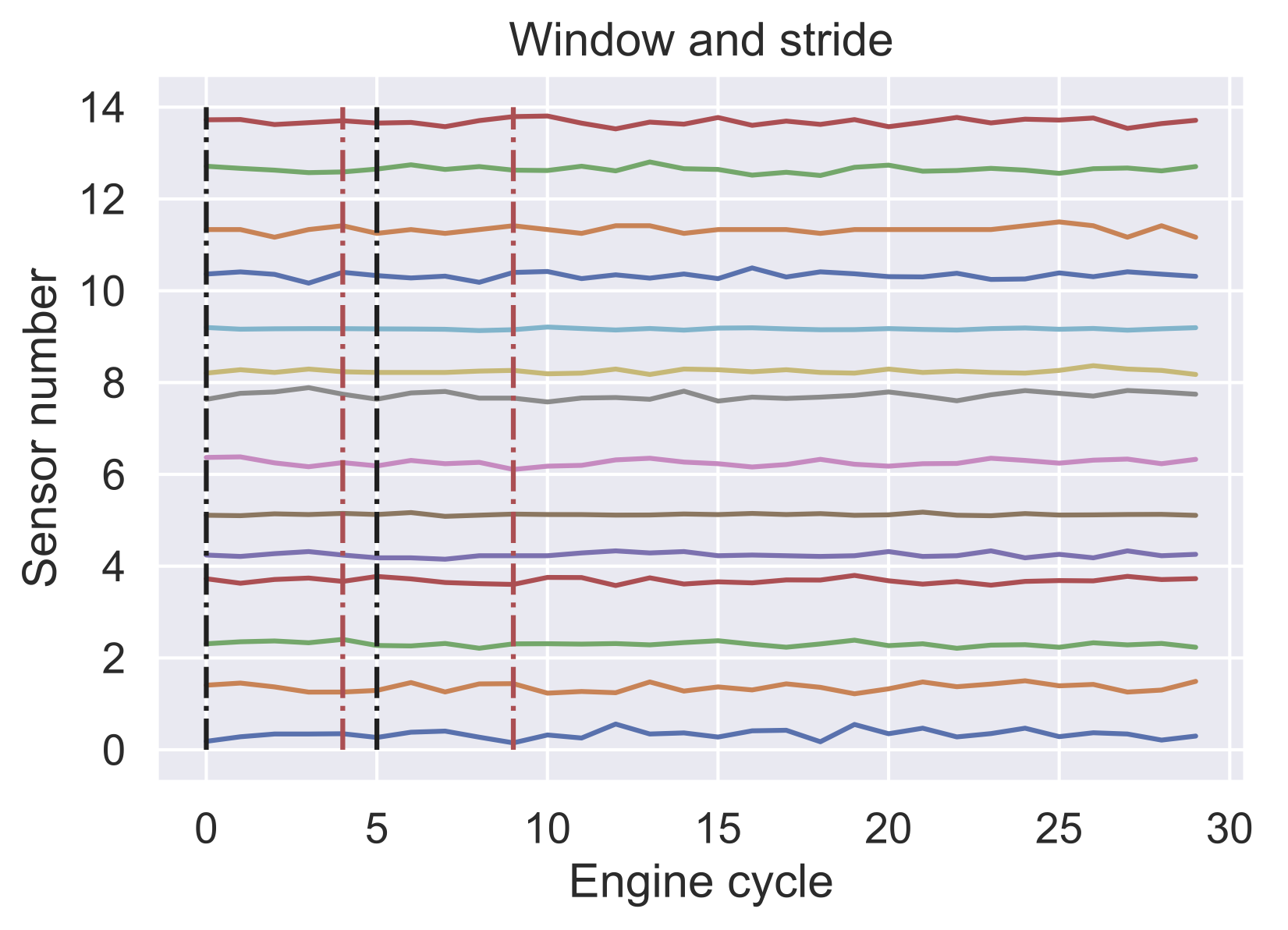}
\caption{Window size and stride example.}
\label{fig:window_stride_arrows}
\end{center}
\end{figure}

Figure \ref{fig:window_stride_arrows} shows an example of how to form a sample vector, in this example $s=14$, $n_w=5$ and $n_s=4$. Each one of the plotted lines denotes the readings for each of the fourteen chosen sensors. The dashed vertical lines (black lines) represent the size of the window, in this case we depict a window size of $5$ cycles. For the next window (red dashed lines) the time window is advanced by a stride of $4$ cycles, note that some sensor readings may be overlapped for different moving windows. For every window, the sensor readings are appended one after another to form a vector of $14*5$ features, for this specific case the unrolled vector will be of $70$ features.

The use of a strided time window allows for the regressor to take advantage not only of the previous information, but also to control the ratio at which the algorithm is fed with new information. With the usual time window approach, only one point is updated for every new time window. The strided time window considered in this study allows for updating more than one point at the time for the algorithm to make use of the new information with less iterations. Our choice of the strided time window is inspired by the use of strided convolutions in  Convolutional Neural Networks \cite{Kong2017}. Further studies of the impact of  the stride on the prediction should be done in the future.

\subsubsection{Piecewise Linear Degradation Model}

\begin{figure}[H]
\begin{center}
\includegraphics[scale=0.7]{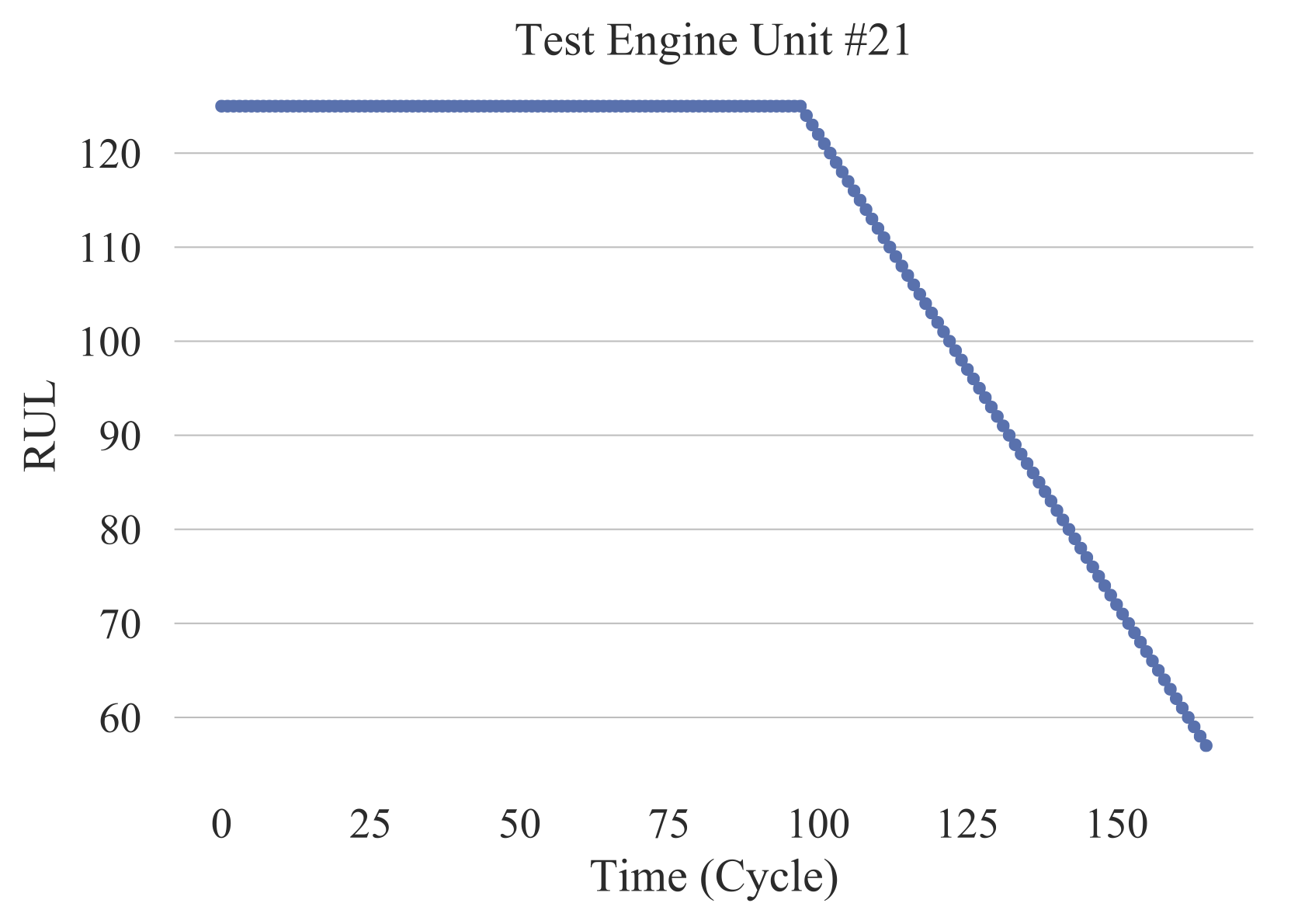}
\caption{Piece-wise linear degradation for RUL.}
\label{FigRULinear}
\end{center}
\end{figure}

Different from common regression problems, the desired output value of the input data is difficult to determine for a RUL problem. It is usually impossible to evaluate the precise health condition and estimate the RUL of the system at each time step without an accurate physics based model. For this popular dataset, a piece-wise linear degradation model has been proposed in \cite{Ramasso2014}. The model assumes that the engines have a constant RUL label in the early cycles, and then the RUL starts degrading linearly until it reaches 0 as shown in Figure \ref{FigRULinear}. The piece-wise linear degradation assumption is used in this work. We denote the value of the RUL in the early cycles as $R_{e}$. Initially,  $R_{e}$ is randomly chosen between 90 and 140  cycles which is a reasonable range of values for this particular application. When the difference between the cycle count in the time window and the terminating cycle 
of the training data is less than the initial value of $R_{e}$, $R_{e}$ begins the linear descent toward the terminating cycle.

\subsection{Optimal Data Parameters}

\label{sec:otimal_data_params}

As mentioned in the previous sections the choice of the data-related parameters $\mathbf{v}$ has a large impact on the performance of the regressor. In this section, we present a method  for picking the optimal combination of the data-related parameters $n_{w}$, $n_{s}$ and $R_{e}$ while being computationally efficient.

Vector $\mathbf{v} = (n_{w}, n_{s}, R_{e})$ components specific to the C-MAPSS dataset are bounded such that $n_{w} \in\left[  1, b\right]  $, $n_{s} \in\left[  1, 10\right]  $, and 
$R_{e} \in\left[  90, 140 \right] $, where all the variables are integer. The value of $b$ is different for different subsets of the data, Table \ref{table:b_values} shows the different values of $b$ for each subset.

\begin{table}[H]
\begin{center}
\begin{tabular}
[c]{c|cccc}\hline
& FD001 & FD002 & FD003 & FD004\\\hline
$b$ & 30 & 20 & 30 & 18\\\hline
\end{tabular}
\caption{Allowed values for $b$ per subset.}
\label{table:b_values}
\end{center}
\end{table}

Let $\mathbf{X}(\mathbf{v})$ be the training/cross-validate/test sets parametrized by $\mathbf{v}$ and $\mathbf{\hat{y}}(\mathbf{v}) = \phi(\mathbf{X}(\mathbf{v}); \boldsymbol{\theta})$ be the predicted RUL values of our model $\phi(.)$ with $\mathbf{X}(\mathbf{v})$. Thus, $E_{\scriptscriptstyle RMS}$ depends directly on the choice of $\mathbf{v}$ since $\phi(.)$ is fixed (See Section \ref{sec:rul_metrics}). Here we propose to solve the following optimization problem

\begin{equation}
\underset{\mathbf{v} \in\mathbb{Z}^{3}}{\mathrm{min} \; \;} E_{\scriptscriptstyle RMS}(\mathbf{v}).
\label{eq:optimization_problem}%
\end{equation}

The problem to find optimal data-related parameters has no analytic descriptions. Therefore, no gradient information is available. An evolutionary algorithm is the natural choice for this optimization problem. Nevertheless, since the computation of the error $E_{\scriptscriptstyle RMS}(\mathbf{v})$ requires re-training $\phi(.)$ an strategy to make the optimization process computationally efficient must be devised.

\subsubsection{True Optimal Data Parameters}

The finite size of C-MAPSS dataset and finite search space  of $\mathbf{v}$ allow an exhaustive search to be performed in order to find the true optimal data-related parameters. We would like to emphasize that although exhaustive search is possible for the C-MAPSS dataset, it is in no way a possibility in a more general setting. Nevertheless, the possibility to perform exhaustive search on the C-MAPSS dataset can be exploited to demonstrate the accuracy of the chosen EA and of the framework overall. In the following studies, we use the results and computational efforts of the exhaustive search as benchmarks to examine the accuracy and efficiency of the proposed approach.

We should note that the subsets of the data FD001 and FD003 have similar features and that the subsets FD002 and FD004 have similar features. Because of this, we have decided to just optimize the data-related parameters by considering the subsets FD001 and FD002 only. An exhaustive search is performed to find the true optimal values for $\mathbf{v}$.  The MLP is only trained for $20$ epochs. Table \ref{table:true_optimal_data_params} shows the optimal as well as the worst combinations of data-related parameters and the total number of function evaluations used by the exhaustive search. It is important to notice that for this experiment the window size is limited to be larger than or equal to $15$.

\begin{table}[H]
\begin{center}
\begin{tabular}
[c]{l|crcrr}\hline
Dataset & argmin $\mathbf{v}$ & min $E_{\scriptscriptstyle RMS}(\mathbf{v})$ & argmax $\mathbf{v}$ & max $E_{\scriptscriptstyle RMS}(\mathbf{v})$ & Function evals.\\\hline
FD001 & $\left[  24,1,127\right]  $ & \multicolumn{1}{c}{$15.11$} & $\left[25,10,94\right]  $ & \multicolumn{1}{c}{$85.19$} & \multicolumn{1}{c}{8160}\\
FD002 & $\left[  16,1,138\right]  $ & \multicolumn{1}{c}{$30.93$} & $\left[17,10,99\right]  $ & \multicolumn{1}{c}{$59.78$} & \multicolumn{1}{c}{3060}\\\hline
\end{tabular}
\caption{Exhaustive search results for subsets FD001 and F002.}
\label{table:true_optimal_data_params}
\end{center}
\end{table}

Numerical experiments seem to suggest that, at least for CMAPSS dataset, the window size plays a big role in terms of the meaningful information used for prediction by the MLP. It also reflects the history-dependent nature of the aircraft engine degradation process. Furthermore, overlapping in the generated time windows seems to benefit the generated sequences of sensors.

\subsubsection{Evolutionary Algorithm for Optimal Data Parameters}

\label{sec:ea_optimization_process}

Evolutionary algorithms (EAs) are a family of methods for optimization problems. The methods do not make any assumptions about the problem, treating it as a black box that merely provides a measure of quality given a candidate solution. Furthermore, EAs do not require the gradient when searching for optimal solutions, making them very suitable for applications such as neural networks.

For the current application, the differential evolution (DE) method is chosen as the optimization algorithm \cite{Storn1997}. Though other meta-heuristic algorithms may also be suitable for this application, the DE has established as one of the most reliable, robust and easy to use EAs. Furthermore, a ready to use Python implementation is available through the scipy package \cite{scipy}. Although the DE method does not have special operators for treating integer variables, a very simple modification to the algorithm, i.e. rounding every component of a candidate solution to its nearest integer, is used for this work.

As mentioned earlier, evolutionary algorithms such as the DE use several function evaluations when searching for the optimal solutions. It is important to consider that, for this application, one function evaluation requires retraining the neural network from scratch. This is not a desirable scenario, as obtaining the optimal data-related parameters would entail an extensive computational effort. Instead of running the DE for several iterations and with a large population size, we propose to run it just for $30$ iterations, i.e. the generations in the literature of evolutionary computation, with a population size of $12$, which seems reasonable given the size of the search space of $\mathbf{v}$.

During the optimization, the MLP is trained for only $20$ epochs. The small number of epochs of training the MLP is reasonable in this case because a small batch of data is used in the training, because we only look for the trend of the scoring indicators. Furthermore, it is common to observe that the parameters leading to lower score values in the early stages of the training are more likely to provide better performance after more epochs of training. The settings of the DE algorithm to find the optimal data-related parameters are listed in Table \ref{table:de_hyperparams}.

\begin{table}[H]
\begin{center}
\begin{tabular}[c]{llll}\hline
Population Size & Generations & Strategy & MLP epochs\\\hline
\multicolumn{1}{c}{12} & \multicolumn{1}{c}{30} & \multicolumn{1}{c}{Best1Bin \cite{Engelbrecht2007}}
& \multicolumn{1}{c}{20}\\\hline
\end{tabular}
\caption{Differential evolution hyper-parameters.}
\label{table:de_hyperparams}
\end{center}
\end{table}

The optimal data-related parameters for the subsets FD001 and FD002 found by the DE algorithm are listed in Table \ref{table:optimal_data_params}. As can be observed, the results are in fact very close to the true optimal ones in Table \ref{table:true_optimal_data_params} for both the subsets of the data. The computational effort is reduced by one order of magnitude when using the DE method as compared to the exhaustive search for the true optimal parameters. From the results in Table \ref{table:optimal_data_params}, it can be observed that the maximum allowable time window is always preferred while, on the other hand, small window strides yield better results. For the case of early RUL, it can be observed that larger values of $R_{e}$ are favored.

\begin{table}[H]
\begin{center}
\begin{tabular}[c]{l|crr}\hline
Dataset & argmin $\mathbf{v}$ & min $E_{\scriptscriptstyle RMS}(\mathbf{v})$ & Function evals.\\\hline
FD001 & $\left[  24,1,129\right]  $ & \multicolumn{1}{c}{$15.24$} & \multicolumn{1}{c}{372}\\
FD002 & $\left[  17,1,139\right]  $ & \multicolumn{1}{c}{$30.95$} & \multicolumn{1}{c}{372}\\\hline
\end{tabular}
\caption{Data-related parameters for each subset obtained with differential evolution.}
\label{table:optimal_data_params}
\end{center}
\end{table}

\subsection{The Estimation Algorithm}

Having described the major building blocks of the proposed method, we now introduce the complete framework in the form of Algorithm \ref{alg:rul_framework}.

\begin{algorithm}[!htb]
\SetAlgoLined
\KwData{Training/testing data $\mathbf{X} \in \mathbb{R}^{m_w \times s*n_w} $}
\SetKwInOut{Input}{Input}\SetKwInOut{Output}{Output}

\Input{Initial set of data-related parameters $\mathbf{v} \in \mathbb{Z}^3$, , training labels $\mathbf{y} \in \mathbb{Z}^m$ and number of training epochs for each evaluation of $\phi(\mathbf{v})$.}
\Output{Optimal set of data-related parameters $\mathbf{v}^* \in \mathbb{Z}^3$.}

\BlankLine

Choose regressor architecture (ANN, SVM, linear/logistic regression, etc).\\
Define $\phi(\mathbf{v})$ as in Section \ref{sec:otimal_data_params}.\\
Optimize $\phi(\mathbf{v})$, by means of an evolutionary algorithm, using the proposed guidelines from Section \ref{sec:ea_optimization_process}.\\
Use $\mathbf{v}^*$ to train the regressor for as many epochs as needed.

\caption{ANN-EA RUL estimation framework.}
\label{alg:rul_framework}
\end{algorithm}

\section{Evaluation of the Proposed Method}

\label{sec:rul_eval}

\subsection{Experimental settings}

In this section, we evaluate the performance of the proposed method. The architecture of the MLP is described in Table \ref{table:proposed_nn}. We define an experiment as the process of training the MLP on any of the subsets (FD001 to FD004) and evaluate its performance using the subset's test set. The combinations of the optimal window size $n_{w}$, window stride $n_{s}$ and early RUL $R_{e}$ are presented in Table \ref{table:data_params_de}. We perform $10$ different experiments for each data subset, the MLP is trained for 200 epochs using the train set for the corresponding data subset and evaluated using the subset's test set. The results for each dataset subset are averaged and presented in Table \ref{table:results_ann_de}. Furthermore, the best model is saved and later used to generate the results presented in Section \ref{sec:experimental_results}. All of the experiments were run using the Keras/Tensorflow framework, an NVIDIA GeForce 1080Ti GPU was used to speed up the training process.

\begin{table}[H]
\begin{center}
\begin{tabular}[c]{l r r r c}\hline
Dataset & $n_{w}$ & $n_{s}$ & $R_{e}$ & Input size (neurons) \\\hline
FD001 & 24 & 1 & 129 & 336\\
FD002 & 17 & 1 & 139 & 238\\
FD003 & 24 & 1 & 129 & 336\\
FD004 & 17 & 1 & 139 & 238\\\hline
\end{tabular}
\caption{Data-related parameters for each subset as obtained by DE.}
\label{table:data_params_de}
\end{center}
\end{table}

\subsection{Experimental results}
\label{sec:experimental_results}

The obtained results for $\phi(\mathbf{v})$ using the above setting are presented in Table \ref{table:results_ann_de}. Notice that the performances obtained for datasets FD001 and FD002 are improved as compared with the results in Table \ref{table:optimal_data_params}. This is due to the fact that the MLP is trained for more epochs, thus obtaining better results.

\begin{table}[H]
\begin{center}
\begin{tabular}[c]{l|cccc|cccc}\hline
& \multicolumn{4}{|c}{$E_{\scriptscriptstyle RMS}$} & \multicolumn{4}{|c}{$E_{\scriptscriptstyle RH}$}\\
Data Subset & min & max & avg & STD & min & max & avg & STD\\\hline
FD001 & 14.24 & 14.57 & 14.39 & 0.11 & 3.25 & 3.58 & 3.37 & 0.11\\
FD002 & 28.90 & 29.23 & 29.09 & 0.11 & 45.99 & 53.90 & 50.69 & 2.17\\
FD003 & 14.74 & 16.18 & 15.42 & 0.50 & 4.36 & 6.85 & 5.33 & 0.95\\
FD004 & 33.25 & 35.10 & 34.74 & 0.53 & 58.52 & 78.62 & 74.77 & 5.88\\\hline
\end{tabular}
\caption{Scores for each dataset using the data-related parameters obtained by DE.}
\label{table:results_ann_de}
\end{center}
\end{table}

\begin{figure}[H]
\begin{center}
\includegraphics[scale=0.7]{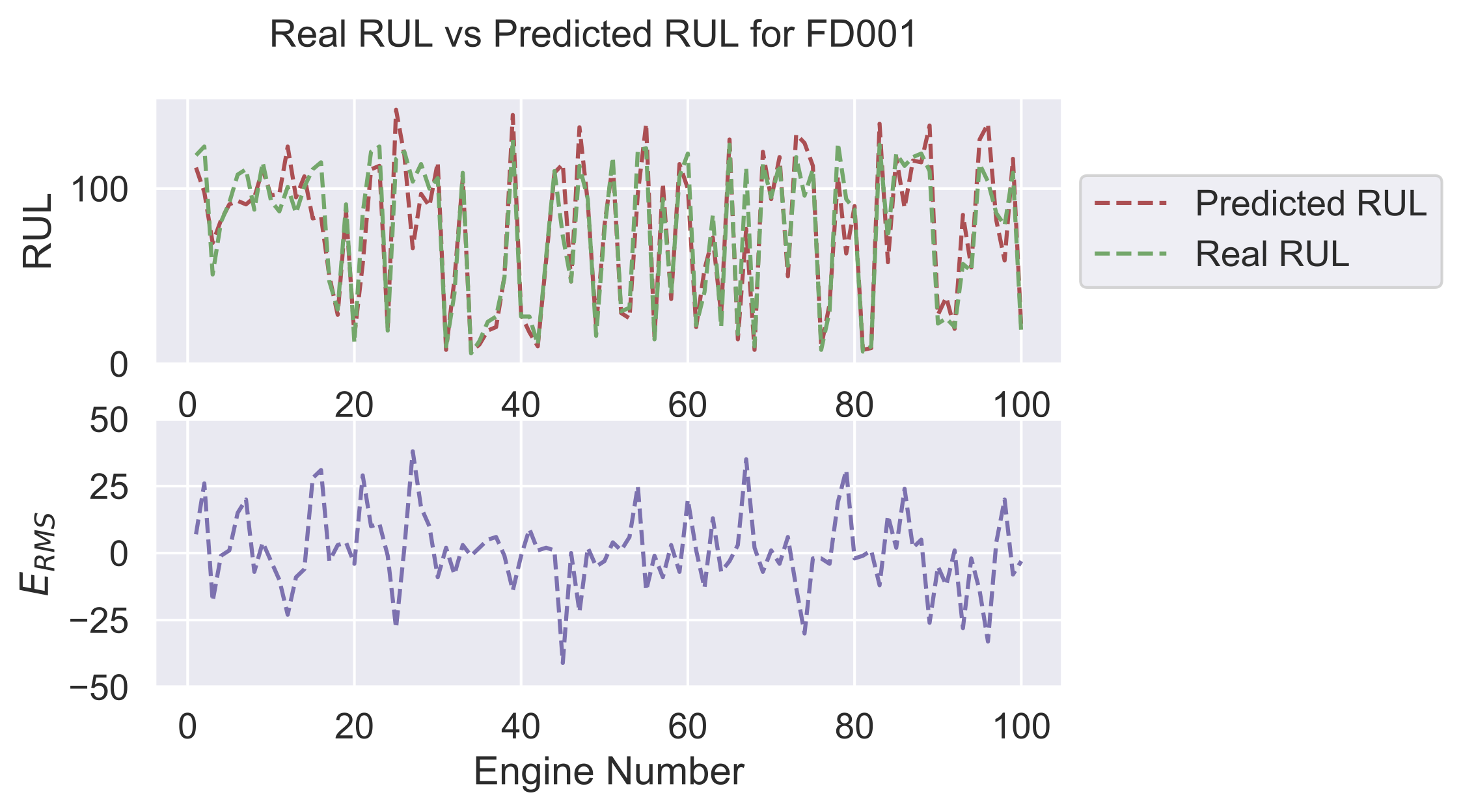}
\caption{Comparison of predicted RUL values vs real RUL values for dataset FD001}
\label{Fig:rul_plots_dataset_1}
\end{center}
\end{figure}

\begin{figure}[H]
\begin{center}
\includegraphics[scale=0.7]{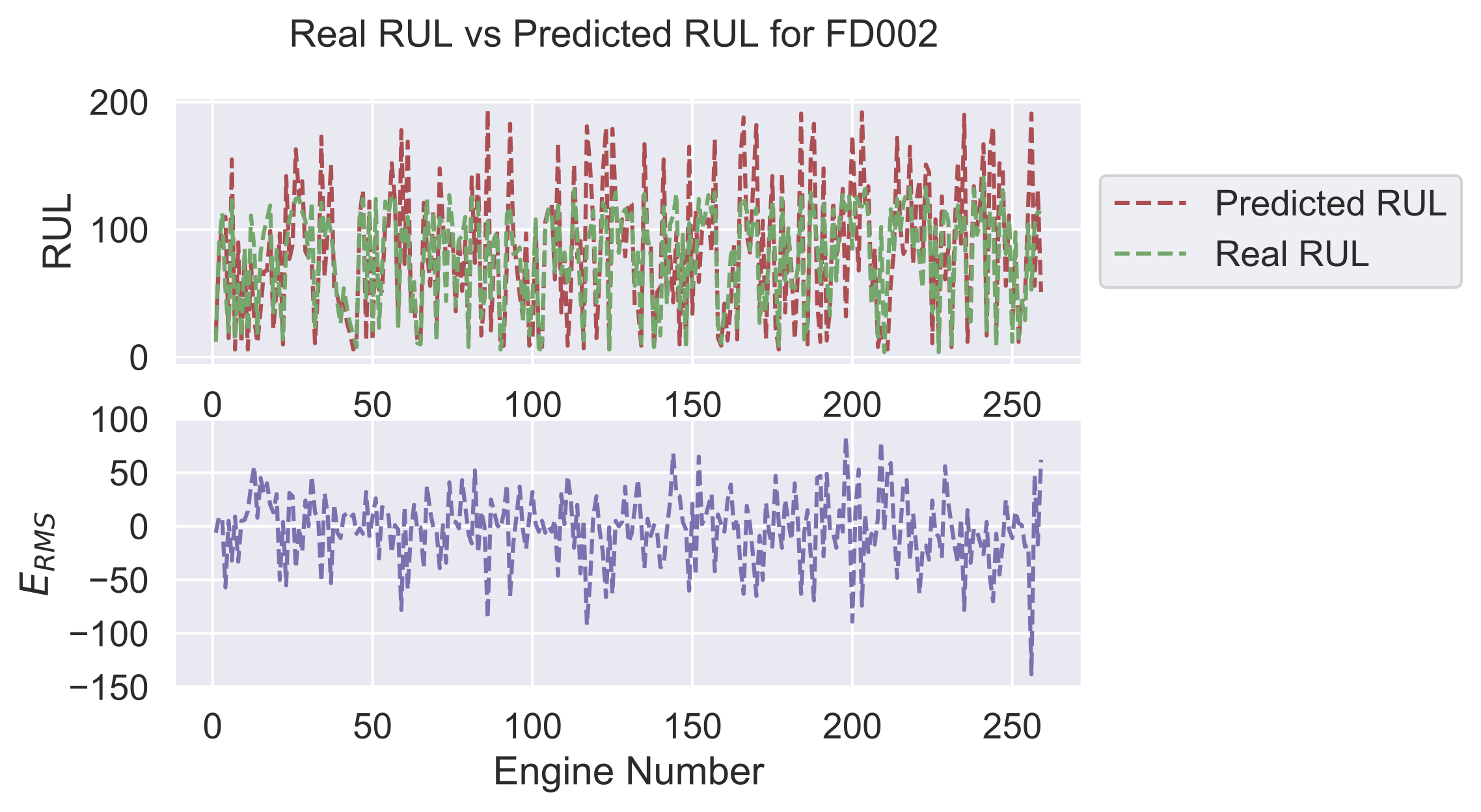}
\caption{Comparison of predicted RUL values vs real RUL values for dataset FD002}
\label{Fig:rul_plots_dataset_2}
\end{center}
\end{figure}

\begin{figure}[H]
\begin{center}
\includegraphics[scale=0.7]{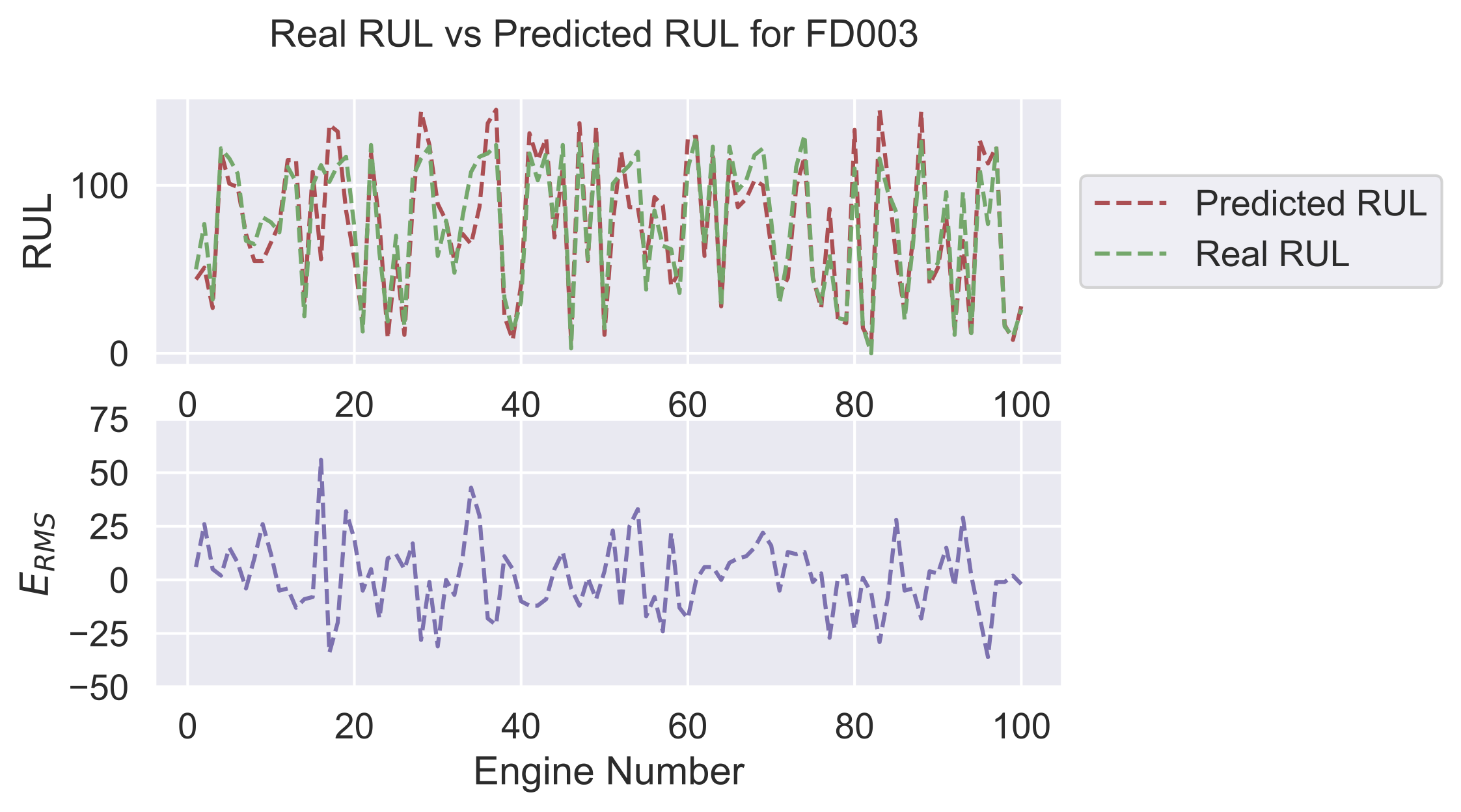}
\caption{Comparison of predicted RUL values vs real RUL values for dataset FD003}
\label{Fig:rul_plots_dataset_3}
\end{center}
\end{figure}

\begin{figure}[H]
\begin{center}
\includegraphics[scale=0.7]{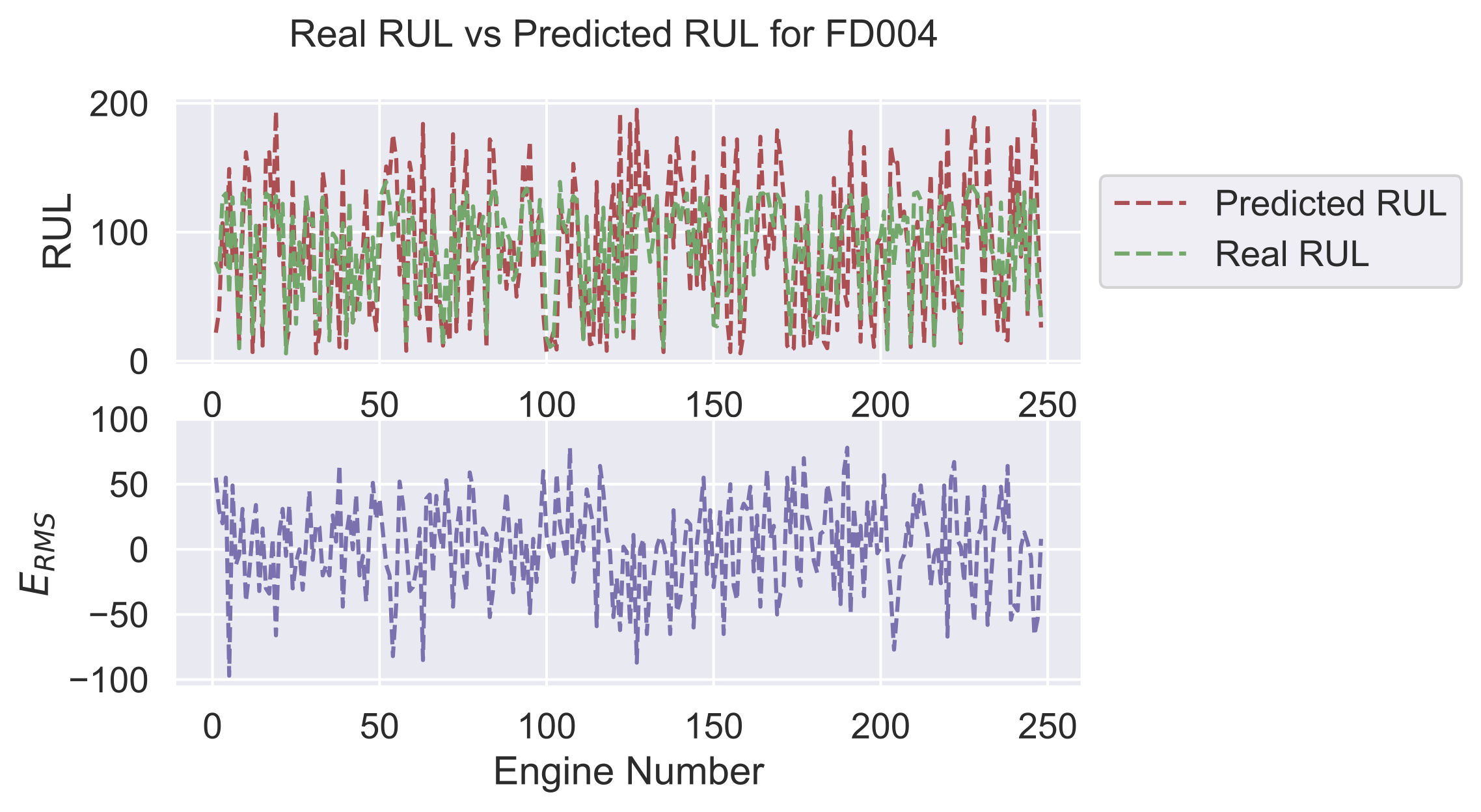}
\caption{Comparison of predicted RUL values vs real RUL values for dataset FD004}
\label{Fig:rul_plots_dataset_4}
\end{center}
\end{figure}

We now compare the predicted RUL values versus the real RUL values for each of the datasets. For Figures \ref{Fig:rul_plots_dataset_1} to \ref{Fig:rul_plots_dataset_4} we plot on the top sub-figure the predicted RUL values (red lines) vs the real RUL values (green lines) while on the bottom sub-figure we plot the error between real RUL and predicted RUL.

Figure \ref{Fig:rul_plots_dataset_1} shows the comparison for subset FD001, it can be observed that the predicted RUL values closely follow the real RUL values with the exception of a pair of engines. The error remains small for most of the engines. Meanwhile, for FD002 it can be observed in Figure \ref{Fig:rul_plots_dataset_2} that the regressor overshoots several of the RUL predictions, especially in the positive spectrum. That is, the method predicts a RUL when in reality the real RUL is less than the predicted value. This is more evident in the second subplot where the maximum error is $138$ at the magenta peak in the leftmost part of the plot.

Figure \ref{Fig:rul_plots_dataset_3} shows that for FD003 the predictions follow closely the real RUL values. The behavior for FD004 is similar to FD002 as depicted in Figure \ref{Fig:rul_plots_dataset_4}, with most of the error such that the predictions of the RUL are larger than the real value.

\pagebreak

\pagebreak

\begin{figure}[H]
\begin{center}
\includegraphics[scale=0.6]{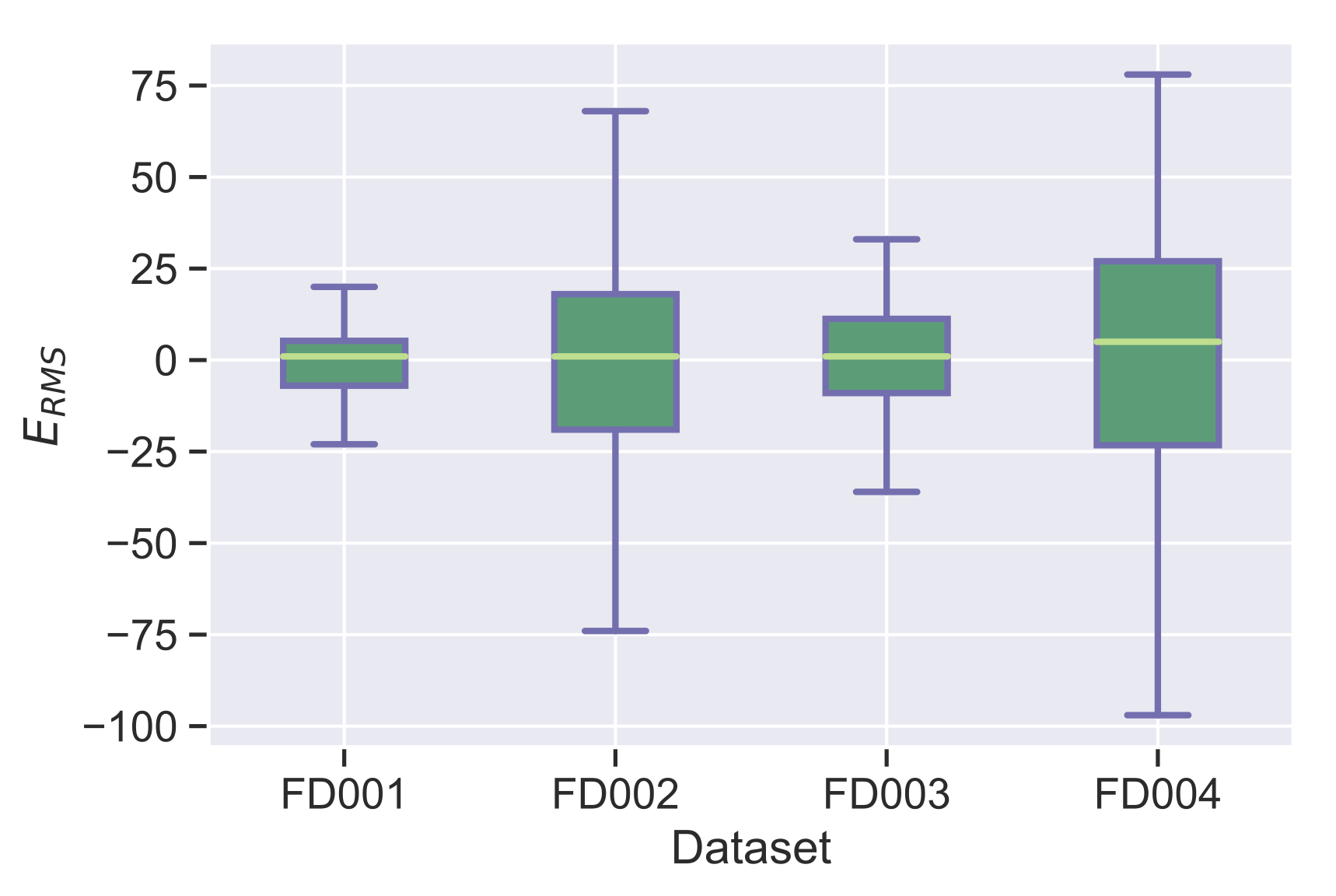}
\caption{Prediction error of the MLP for each dataset.}
\label{Fig:rul_plots_boxplot}
\end{center}
\end{figure}

Finally, Figure \ref{Fig:rul_plots_boxplot} shows that most of the RUL predictions by the method are highly accurate. In the case of FD001, the predictions of $50\%$ of the engines is smaller than $6$ cycles. In the case of FD002, the predictions are acceptable with the error of first quartile being lower than $6$ cycles and the error for $50\%$ of the engines being less than $19$ cycles. The cases for FD003 and FD004 are similar to FD001 and FD002, respectively.

Two conclusions can be drawn from the previous discussions. First, it can be observed that the number of operating conditions has a larger impact on the complexity of the data than the number of fault modes. This is because the subsets FD002 and FD004 exhibit larger errors than the subsets FD001 and FD003, although the general trend of prediction errors is similar among those two groups of data sets. Second, for the subsets FD002 and FD004, most of the error is related to the predictions larger than the real RUL values. Another observation is that larger window sizes usually lead to better predictions. This may be related to the history-dependent nature of the physical problem.

\subsection{Comparison with other approaches}

The performance of the proposed method is also compared against other state-of-the-art methods. The methods chosen for the comparison obey to two criteria: 1) that the method used is a machine learning (or statistical) method and 2) that the method is recent (6 years old at most). Most of the methods chosen here have only reported results on the test set FD001 in terms of $E_{rms}$. The results are shown in Table \ref{table:results_comparison}. The $E_{rms}$ value of the proposed method in Table \ref{table:results_comparison} is the mean value of 10 independent runs. The values of other methods are identical to those reported in their respective original papers.

\begin{table}[H]
\begin{center}
\begin{tabular}[c]{l|r}\hline
Method & $E_{rms}$\\\hline
ESN trained by Kalman Filter \cite{Peng2012} & 63.45\\
Support Vector Machine Classifier \cite{Louen2013} & 29.82\\
Time Window Neural Network \cite{Lim2016} & 15.16\\
Multi-objective deep belief networks ensemble \cite{Zhang2016} & 15.04\\
Deep Convolutional Neural Network \cite{Babu2016} & 18.45\\
\textbf{Proposed method with $n_{w}=30$, $n_{s}=1$ and $R_{e}=128$} & \textbf{14.39}\\
Modified approach with classified sub-models of the ESN \cite{Peng2012}. & 7.021\\
Only 80 of 100 engines predicted & \\
Deep CNN with time window \cite{Li2018}. & 13.32\\
RNN-Encoder-Decoder \cite{Malhorta2016}. & 12.93\\
\hline
\end{tabular}
\caption{Performance comparison of the proposed method and the latest related papers on the C-MAPSS dataset.}
\label{table:results_comparison}
\end{center}
\end{table}

From the comparison studies, we can conclude that the proposed method performs better than the majority of the chosen methods when taking into consideration the whole dataset FD001. Two existing methods come close to the performance of the proposed approach here, namely the time window ANN \cite{Lim2016} and the Networks Ensemble \cite{Zhang2016}.
While the performance of these two methods comes close to the results of the proposed method, our method is computationally more efficient. We believe that the use of the time window  with a proper size makes the difference. Notice that three of the methods presented in Table \ref{table:results_comparison} perform better than the proposed method, namely, the Convolutional Neural Network \cite{Li2018}, the RNN-Encoder-Decoder \cite{Malhorta2016} and the modified ESN \cite{Peng2012}. Nevertheless in the case of \cite{Peng2012}, the method can only predict 80 out of the 100 total engines while the deep CNN \cite{Li2018} and RNN \cite{Malhorta2016} approaches are much more computationally expensive than the MLP used in this work. In the case of the method proposed in \cite{Li2018}, the neural network model has four layers of convolutions and two more layers of fully connected layers. On the other hand, the RNN-Encoder-Decoder \cite{Malhorta2016} makes use of a much more complicated scheme of two RNNs, one for encoding the sequences and one for decoding them. While specialized libraries such as TensorFlow or Keras make RNNs easier to use, they still remain up to date as some of the most computationally expensive architecture to train given their sequential nature. Finally, we would like to emphasize that the used MLP for this approach is one of the simplest ones in the reviewed literature. Furthermore, the framework proposed is simple to understand and implement, robust, generic and light-weight. These are the features important to highlight when comparing the proposed method against other state-of-the-art approaches.

\section{Conclusions}

\label{sec:conclusions}

We have presented a novel framework for predicting the RUL of mechanical
components. While the method has been tested on the jet-engine dataset
C-MAPSS, the method is general enough such that it can be applied to other similar
systems. The framework makes use of a strided moving time window to generate
the training and test records. A shallow MLP to make the predictions of the RUL
has been found to be sufficient for the C-MAPSS dataset. The evolutionary
algorithm DE needs to be run just once to find the best data-related
parameters that optimize the scoring functions. The resulting model of the application of the framework presented in this
paper demonstrated to be accurate and computationally
efficient, specially when applied to large datasets in real applications. The compactness of the resulting model makes it suitable for applications that have
limited computational resources such as embedded systems. It is important to note that the framework itself does involve some computations, nevertheless such computations are done off-line. Furthermore, the
comparison with other state-of-the-art methods has shown that the proposed
method is the best overall performer.

Two major features of the proposed framework are its generality and
scalability. While for this study, specific regressors and evolutionary
algorithms are chosen, many other combinations are possible and may be more
suitable for different applications. Furthermore, the framework can, in
principle, be used for model-construction, i.e. generating the best possible
neural network architecture tailored to a specific application. Thus, future work will consider extensions to the framework to make it applicable to tasks such as model selection. An analysis of the influence of the window stride parameter $n_s$ will also be considered for future work.

\section*{Acknowledgement}
The authors acknowledge the funding from Conacyt Project No. 285599 and a grant (11572215) from the National Natural Science Foundation of China.

\section*{References}

\bibliographystyle{elsarticle-num}
\bibliography{reference_rul_paper}
%


\clearpage

\onecolumn%

\bigskip%

%

\clearpage

\clearpage

\appendix

\section{Tested Neural Network Architectures}
\label{appendix:tested_architectures}

\setcounter{table}{0}
\bigskip

In this appendix we present the tested neural network architectures. Each table represents a neural network model. Each row in the table represents a neural network layer while each column describes each one of the key parameters of the layer such as the type of layer, number of neurons in the layer, activation function of the layer and whether regularization is used, where L1 denotes the L1 regularization factor and L2 denotes the L2 regularization factor, the order in which the layers are appended from the table is top-bottom.

\begin{table}[H]
\centering
\caption{Proposed neural network architecture 1.}%
\begin{tabular}
[c]{llll}\hline
Layer & Neurons & Activation & Additional Information\\\hline
Fully connected & 20 & ReLU & L1 = 0.1, L2 = 0.2\\
Fully connected & 20 & ReLU & L1 = 0.1, L2 = 0.2\\
Fully connected & 1 & Linear & L1 = 0.1, L2 = 0.2\\\hline
\end{tabular}
\label{table:proposed_nn_1}%
\end{table}

\begin{table}[H]
\centering
\caption{Proposed neural network architecture 2.}%
\begin{tabular}
[c]{llll}\hline
Layer & Neurons & Activation & Additional Information\\\hline
Fully connected & 50 & ReLU & L1 = 0.1, L2 = 0.2\\
Fully connected & 20 & ReLU & L1 = 0.1, L2 = 0.2\\
Fully connected & 1 & Linear & L1 = 0.1, L2 = 0.2\\\hline
\end{tabular}
\label{table:proposed_nn_2}%
\end{table}

\begin{table}[H]
\centering
\caption{Proposed neural network architecture 3.}%
\begin{tabular}
[c]{llll}\hline
Layer & Neurons & Activation & Additional Information\\\hline
Fully connected & 100 & ReLU & L1 = 0.1, L2 = 0.2\\
Fully connected & 50 & ReLU & L1 = 0.1, L2 = 0.2\\
Fully connected & 1 & Linear & L1 = 0.1, L2 = 0.2\\\hline
\end{tabular}
\label{table:proposed_nn_3}%
\end{table}

\begin{table}[H]
\centering
\caption{Proposed neural network architecture 4.}%
\begin{tabular}
[c]{llll}\hline
Layer & Neurons & Activation & Additional Information\\\hline
Fully connected & 250 & ReLU & L1 = 0.1, L2 = 0.2\\
Fully connected & 50 & ReLU & L1 = 0.1, L2 = 0.2\\
Fully connected & 1 & Linear & L1 = 0.1, L2 = 0.2\\\hline
\end{tabular}
\label{table:proposed_nn_4}%
\end{table}

\begin{table}[H]
\centering
\caption{Proposed neural network architecture 5.}%
\begin{tabular}
[c]{llll}\hline
Layer & Neurons & Activation & Additional Information\\\hline
Fully connected & 20 & ReLU & L1 = 0.1, L2 = 0.2\\
Fully connected & 1 & Linear & L1 = 0.1, L2 = 0.2\\\hline
\end{tabular}
\label{table:proposed_nn_5}%
\end{table}

\begin{table}[H]
\centering
\caption{Proposed neural network architecture 6.}%
\begin{tabular}
[c]{llll}\hline
Layer & Neurons & Activation & Additional Information\\\hline
Fully connected & 10 & ReLU & L1 = 0.1, L2 = 0.2\\
Fully connected & 1 & Linear & L1 = 0.1, L2 = 0.2\\\hline
\end{tabular}
\label{table:proposed_nn_6}%
\end{table}


\pagebreak

\section{Activation functions definitions}
\label{appendix:activation_definitions}

Here we define some of the activation functions used in our neural network models. For the following definitions assume a vector $\mathbf{x} \in \mathbb{R}^n$.

\subsection{ReLU activation function}

\begin{figure}
[H]
\begin{center}
\includegraphics[scale=0.6]%
{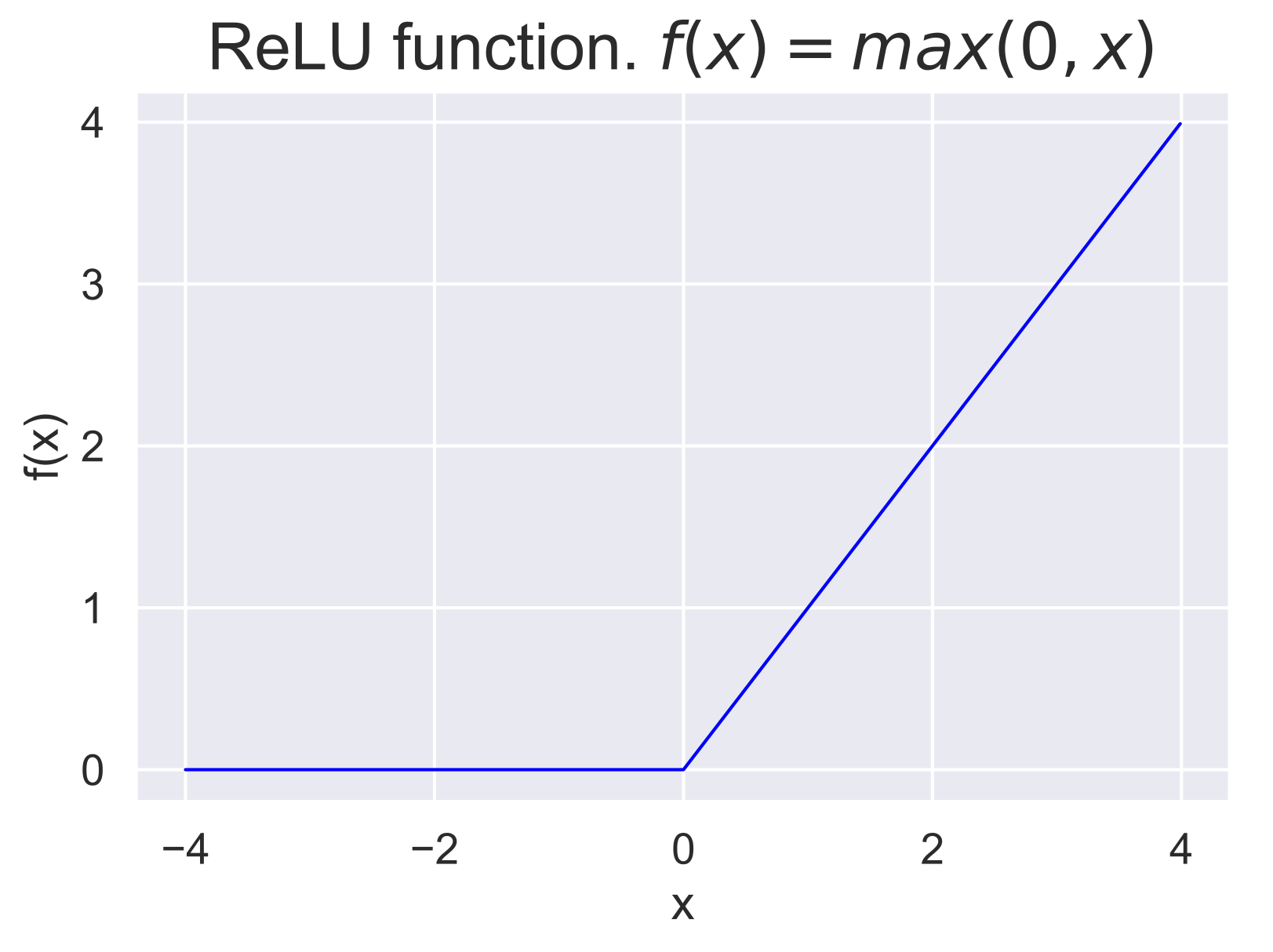}%
\caption{ReLU activation function}%
\label{Fig:relu_function}%
\end{center}
\end{figure}

\subsection{Linear activation function}

\begin{figure}
[H]
\begin{center}
\includegraphics[scale=0.6]%
{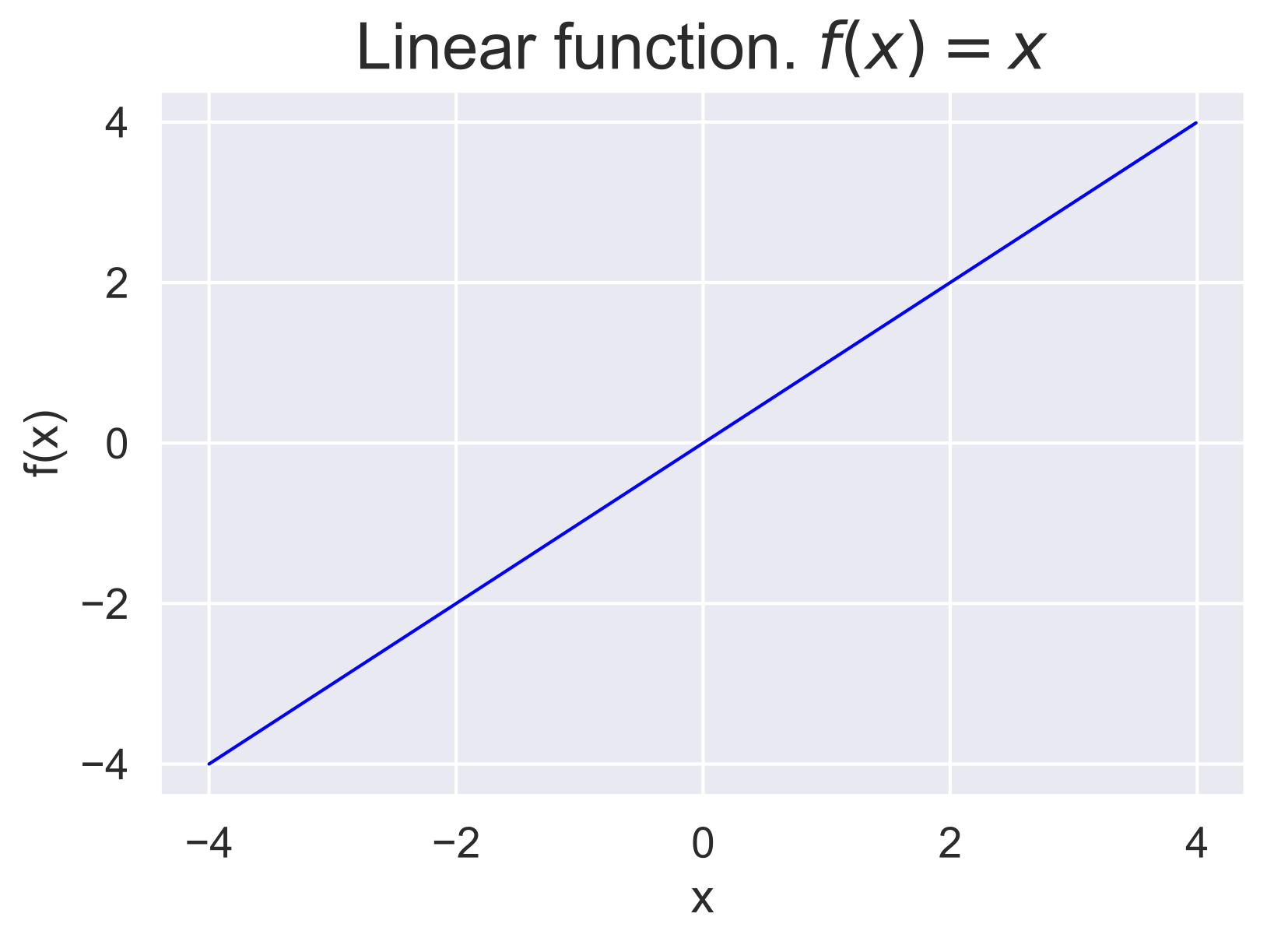}%
\caption{Linear activation function}%
\label{Fig:linear_function}%
\end{center}
\end{figure}

\end{document}